\documentclass[runningheads]{llncs}
\usepackage{header}

\usepackage{graphicx}

\usepackage{times}
\usepackage{latexsym}

\usepackage[T1]{fontenc}

\usepackage[utf8]{inputenc}

\usepackage{microtype}

\usepackage{amsmath,amssymb}
\usepackage{pifont}
\usepackage{algorithmic}
\usepackage{graphicx}
\usepackage{textcomp}
\usepackage{xcolor}
\usepackage{multirow}
\usepackage{tabularx,ragged2e}
\usepackage{tabulary}
\usepackage{adjustbox}
\usepackage{hyperref}
\usepackage{cleveref}
\usepackage[flushleft]{threeparttable}
\usepackage{url}
\usepackage{hhline}
\usepackage{multicol}
\usepackage{booktabs}
\usepackage{footmisc}
\usepackage{graphicx}
\usepackage{arydshln}
\usepackage{tabularx}
\usepackage{color, colortbl}

\makeatletter
\newcommand{\printfnsymbol}[1]{%
  \textsuperscript{\@fnsymbol{#1}}%
}
\makeatother

\definecolor{Gray}{gray}{0.945} %
\definecolor{darkblue}{rgb}{0, 0, 0.5}
\hypersetup{colorlinks=true, citecolor=darkblue, linkcolor=darkblue, urlcolor=darkblue}

\begin{document}

\title{Testing the Generalization of Neural Language Models for COVID-19 Misinformation Detection}

\titlerunning{Testing the Generalization of neural LMs for COVID-19 Misinformation Detection}

\author{Jan Philip Wahle\thanks{Equal contribution.}\inst{1}\orcidID{0000-0002-2116-9767} \and Nischal Ashok\printfnsymbol{1}\inst{2}\orcidID{0000-0002-7022-5948} \and Terry Ruas\inst{1}\orcidID{0000-0002-9440-780X} \and Norman Meuschke\inst{1}\orcidID{0000-0003-4648-8198} \and Tirthankar Ghosal\inst{3}\orcidID{0000-0002-2358-522X} \and Bela Gipp\inst{1}\orcidID{0000-0001-6522-3019}}
%
\authorrunning{ }

\institute{University of Wuppertal, Rainer-Gruenter-Straße, 42119 Wuppertal, Germany \email{last@uni-wuppertal.de} \\ \and Indian Institute of Technology Patna, Bihar-801106, India \\ \email{1801cs33@iitp.ac.in} \\ \and Charles University, Malostranské náměstí 25, 118 00 Praha, Czech Republic  \\ \email{ghosal@ufal.mff.cuni.cz}}

\maketitle

\thispagestyle{firststyle}

\begin{abstract}
A drastic rise in potentially life-threatening misinformation has been a by-product of the COVID-19 pandemic. Computational support to identify false information within the massive body of data on the topic is crucial to prevent harm. Researchers proposed many methods for flagging online misinformation related to COVID-19. However, these methods predominantly target specific content types (e.g., news) or platforms (e.g., Twitter). The methods' capabilities to generalize were largely unclear so far. We evaluate fifteen Transformer-based models on five COVID-19 misinformation datasets that include social media posts, news articles, and scientific papers to fill this gap. We show tokenizers and models tailored to COVID-19 data do not provide a significant advantage over general-purpose ones. Our study provides a realistic assessment of models for detecting COVID-19 misinformation. We expect that evaluating a broad spectrum of datasets and models will benefit future research in developing misinformation detection systems.

\end{abstract}
\section{Introduction}
The COVID-19 pandemic has claimed more than four million lives by the time of writing this paper, and the number of infections remains high\footnote{\url{https://coronavirus.jhu.edu/map.html}}. The behavior of individuals strongly affects the risk of infection. In turn, the quality of information individuals receive strongly influences their actions~\cite{PennycookMZL20,HaleAGK21}. The novelty and rapid global spread of the SARS-CoV-2 virus has also led to countless life-threatening incidences of misinformation spread on the topic. Controlling COVID-19 and combating possible future pandemics early, requires reducing misinformation and increasing the distribution of facts on the subject~\cite{CinelliQGV20,Zarocostas20}.

Researchers worldwide collaborate on automating the detection of false information on COVID-19.\footnote{We collectively refer to fake news, disinformation, and misinformation as false information.} The initiatives build collections of scientific papers, social media posts, and news articles to analyze their content, spread, source, and propagators \cite{WangLCR20,CuiL20,MutluOJT20}.

Natural Language Processing (NLP) research has extensively studied options to automate the identification of fake news~\cite{ShuSWT17}, primarily by applying recent language models. Researchers proposed adaptions of well-known Transformer models, such as COVID-Twitter-BERT \cite{MullerSK20}, to identify false information on COVID-19 from specific sources. However, most prior studies analyze specific content types (e.g., news) or platforms (e.g., Twitter). These limitations prevent reliable conclusions regarding the generalization of the proposed language models.
  
To fill this gap, we apply 15 Transformer models to five COVID-19 misinformation tasks. 
We compare Transformer models optimized on COVID-19 datasets to state-of-the-art neural language models. We exhaustively apply models to different tasks to test their generalization on unknown sources.
The code to reproduce our experiments,\footnote{\url{https://github.com/ag-gipp/iConference22_COVID_misinformation}} and the datasets used are publicly available.

\section{Related Work}

The same way word2vec~\cite{MikolovSCC13} inspired many models in NLP~\cite{BojanowskiGJM17,RuasGA19,RuasFGd20}, the excellent performance of BERT~\cite{DevlinCLT19}, a Transformer-based model~\cite{VaswaniSPU17}, caused its numerous adaption for language tasks~\cite{YangDYC19,ClarkLLM20,WahleRMF21,WahleRMG21}.
Domain-specific models build on top of Transformers typically outperform their baselines for related tasks~\cite{HowardR18}. For example, SciBERT~\cite{BeltagyLC19a} was pre-trained on scientific documents and typically outperforms BERT for scientific NLP tasks, such as determining document similarity~\cite{OstendorffRBG20}.

Many models for COVID-19 misinformation detection employ domain-specific pre-training to improve their representation.
COVID-Twitter-BERT~\cite{MullerSK20} was pre-trained on 160M tweets and evaluated for sentiment analysis of tweets, e.g., about COVID vaccines. BioClinicalBERT~\cite{AlsentzerMBW19} was trained into clinical narratives to incorporate linguistic characteristics from the biomedical and clinical domains.

Cui et al. \cite{CuiL20} investigated the misinformation detection task by comparing traditional machine learning and deep learning techniques.
Similarly, Zhou et al. \cite{ZhouMFZ20} explored statistical learners, such as SVM, and neural networks to classify news as credible or not.
The results of both studies show deep learning architectures as the most prominent alternatives for the respective datasets.

As papers on COVID-19 are recent, some contributions are only available as pre-trained models. COVID-BERT\footnote{\url{https://tinyurl.com/86cpx6u2}} and COVID-SciBERT\footnote{\url{https://tinyurl.com/9w24pc93}} are pre-trained on the CORD-19 dataset and only available via the Huggingface API. Others, such as COVID-CQ~\cite{MutluOJT20} and CMU-MisCov19~\cite{MemonC20} are used to either investigate intrinsic details (e.g., how dense misinformed communities are) or to explore the applicability of statistical techniques.

Although related works provide promising approaches to counter misinformation related to COVID-19, none of them explore multiple datasets. Research in many NLP areas already uses diverse benchmarks to compare models~\cite{WangSMH19,WangPNS19}. To the best of our knowledge, our study is the first to systematically test Transformer-based methods on different data sources related to COVID-19.

\section{Methodology}
\label{sec:methodology}

\textbf{Models}. Our study includes 15 Transformer-based models, which are detailed in \Cref{sec:appendix_model_details}. We categorize the models into the following three groups:

\textbf{General-Purpose Baselines}. The first group consists of general-purpose Transformer models without domain-specific training, i.e., BERT~\cite{DevlinCLT19}, RoBERTa~\cite{LiuOGD19}, BART~\cite{LewisLGG19}, DeBERTa~\cite{HeLGC21}. These baselines show how vanilla Transformer-based models perform on the COVID-19 misinformation detection task.

\textbf{Intermediate Pre-Training}. The second group contains models trained on specific content types and domains, i.e., SciBERT~\cite{BeltagyLC19a}, BERTweet~\cite{NguyenVT20}, and BioClinicalBERT~\cite{AlsentzerMBW19}. For example, SciBERT adapts BERT for scientific articles. These models show the effect of intermediate pre-training on specific sources compared to general-purpose training (e.g., whether BERTweet is superior to BERT for misinformation on Twitter). Moreover, we compare the models in this group to language models optimized using intermediate pre-training on COVID-19 data (third group).

\textbf{COVID-19 Intermediate Pre-Training}. The third group comprises models employing an intermediate pre-training stage on COVID-19 data. Due to task-specific pre-training, we expect these models to achieve better results than the models in groups one and two. We include a model that optimizes the pre-training objective on a large Twitter corpus, i.e., CT-BERT~\cite{MullerSK20}, two models trained on the CORD-19 dataset (COVID-BERT\footnote{\url{https://tinyurl.com/86cpx6u2}} and COVID-SciBERT\footnote{\url{https://tinyurl.com/9w24pc93}}), and two popular models from the huggingface API for which the intermediate pre-training sources are not released yet (ClinicalCOVID-BERT\footnote{\url{https://tinyurl.com/kebysw}} and BioCOVID-BERT\footnote{\url{https://tinyurl.com/4xx9vdkm}}). 
We pre-train RoBERTa, BART, and DeBERTa on the CORD-19 dataset to compare them to the models we used as general-purpose baselines.

\noindent
\textbf{Data}.
We compile an evaluation set from six popular datasets for detecting COVID-19 misinformation in social media, news articles, and scientific publications, i.e., CORD-19~\cite{WangLCR20}, CoAID~\cite{CuiL20}, COVID-CQ~\cite{MutluOJT20}, ReCOVery~\cite{ZhouMFZ20}, CMU-MisCov19~\cite{MemonC20}, and COVID19FN.\footnote{\url{https://tinyurl.com/4ne9vtzu}} \Cref{tab:datasets} gives an overview of the datasets and \Cref{sec:appendix_dataset_details} presents more details. For CORD-19, we only use abstracts in the dataset, as they provide an adequate trade-off between size and information density. Additionally, less than 50\% of the articles in CORD-19 are available as full texts. For CoAID and ReCOVery, we only extract news articles to reduce a bias towards Twitter posts in our evaluation. All remaining datasets are used in their original composition.

We use CORD-19 to extend the pre-training of general-purpose models and all other datasets to evaluate the models for a downstream task. CORD-19 consists of scientific articles, while the other datasets primarily contain news articles and Twitter content. We chose different domains for training and evaluation to test the models' generalization capabilities and avoid overlaps between the datasets.

\begin{table*}[!htb]
\caption{An overview of the COVID-related datasets. CORD-19 has no specific \textit{Task} or \textit{Label} as it provides a general collection of documents. $^\dagger$Details on the labels are given in Memon et al. \cite{MemonC20}.} \label{tab:datasets}
\centering
\resizebox{1\textwidth}{!}{
    \begin{tabular}{llllll}\toprule
    \textbf{Corpus} & \textbf{$|$Corpus$|$} &  \textbf{Task} & \textbf{Domain} & \textbf{Source(s)} & \textbf{Label(s)} \\
    \midrule
    
    CORD-19 (\cite{WangLCR20}) & 497\,906 & - & Scientific articles & CZI, PMC, BioRxiv, MedRxiv & - \\
    CoAID (\cite{CuiL20}) & 302\,926 & Misinformation detection & Healthcare misinformation & Twitter,
    news, 
    social media
    & \{\textit{true, false}\}  \\
    ReCOVery (\cite{ZhouMFZ20}) & 142\,849 & Credibility classification & Low information credibility & Twitter, news & \{\textit{reliable, unreliable}\}\\
    COVID-CQ (\cite{MutluOJT20}) & 14\,374 & Efficacy of treatments & Drug treatment & Twitter & \{\textit{neutral, against, for}\} \\
    CMU-MisCov19 (\cite{MemonC20}) & 4\,573 & Communities detection  & Misinformed communities & Twitter & \{\textit{17 labels$^\dagger$}\} \\
    COVID19FN (2020) & 2\,800 & Misinformation detection & Misinformation in news & Poynter &  \{\textit{true, false}\}\\
    
    \bottomrule
    \end{tabular}
}
\end{table*}


\section{Experiments}
\label{sec:experiments}

\textbf{Overview.} Our study includes three experiments.
The first experiment tests how static word embeddings and frozen contextual embeddings perform compared to fine-tuned language models. The second experiment studies whether tokenizers specifically tailored to a COVID-19 vocabulary are superior to general-purpose ones.\footnote{General-purpose refers to the tokenizers released with the pre-trained models.} The third experiment evaluates and compares all 15 Transformer models on the five evaluation datasets.

\textbf{Training \& Evaluation.} To compare general-purpose baselines and COVID-19 intermediate pre-trained models, we perform \textit{pre-training} on the CORD-19 dataset for three models (RoBERTa, BART, and DeBERTa) and use pre-trained configurations for the remaining models (BERT, SciBERT, BioCOVID-BERT, ClinicalCOVID-BERT).
We then \textit{fine-tune} all models 
for each of the five test tasks (COVID-CQ, CoAID,  ReCOVery, CMU-MisCov19, and COVID19FN). We use a split of 80\% and 20\% of the documents in a dataset for training and testing, respectively. This split is the most common configuration for the tested datasets \cite{ZhouMFZ20,MutluOJT20} and is comparable to other studies \cite{CuiL20}. We use 10\% of the train dataset as a hold-out validation set.

\section{Results \& Discussion}

\textbf{Static and Frozen Embeddings}.
\Cref{tab:frozen_weights} compares the classification results of a baseline composed of BiLSTM and GlobalVectors (GloVe) \cite{BojanowskiGJM17} to the frozen embeddings of three Transformer models for the COVID-CQ dataset. The results show no significant difference between GloVe and the frozen models. However, fine-tuning the same three models end-to-end generally increases their performance. Therefore, we choose to fine-tune neural language models for the classification of COVID-19 misinformation.

\textbf{Tokenizer Ablation}.
\Cref{tab:tokenizer_inv} shows the results on COVID-CQ for the best configuration of the models using a standard\footnote{Pre-Trained tokenizer provided by HuggingFace} tokenizer for pre-training and fine-tuning. 
We expected adjusting the tokenizer to the CORD-19 dataset would improve the results, as it adds valuable tokens to the vocabulary, which are often not present in standard tokenizers.
However, using specialized tokenizers decreased the performance. The content in CORD-19 originates from the scientific domain. We hypothesize tweets lack similar token relations, which causes the performance drop on the COVID-CQ dataset.
Therefore, we use the standard tokenizer for our full evaluation experiments.

\textbf{Full Evaluation}.
\Cref{tab:fine_tuned} reports the results of our full evaluation. All results are statistically significant using bootstrap and permutation tests ($p < .05$) \cite{DrorBSR18}. General-purpose baselines achieved the best result for two of the five datasets (BART on CoAID and BART on COVID19FN).
For two datasets (ReCOVery and COVID-CQ), a model we pre-trained on CORD-19 data (COVID-RoBERTa) performed best. 
CT-BERT achieved the best result on CMU-MisCov19, an expected outcome as the datasets consist only of Twitter content. BERTweet, which was also trained on Twitter data, does not achieve better results than general-purpose baselines. We expected a minor drop in performance for BERTweet compared to CT-BERT as the former was not trained on COVID-19 vocabulary, but better a performance than general-purpose models as BERTweet was trained mainly on Twitter data.

\begin{table*}[!htb]
    \begin{minipage}[t]{.47\linewidth}
      \centering
        \footnotesize
        \caption{F1-Macro scores of neural language models and a baseline (BiLSTM+GloVe) for the COVID-CQ dataset. The \textit{static} and \textit{frozen} models use a stacked BiLSTM; \textit{fine-tuned} models were pre-trained on the CORD-19 dataset and fine-tuned for the task.}     \label{tab:frozen_weights}
        \begin{threeparttable}
        \begin{tabular}{clr}\toprule
        \textbf{Type} &\textbf{Models} & \textbf{F1-Macro}      \\ \midrule
        \textit{static} & GloVe    & .71   \\ \hdashline\noalign{\vskip 0.5ex}
        \textit{frozen} & BERT  & .72  \\
        \textit{frozen} & RoBERTa & .70 \\ 
        \textit{frozen} & SciBERT & .68 \\\hdashline\noalign{\vskip 0.5ex}
        \textit{fine-tuned} & BERT  & \textbf{.75}  \\
        \textit{fine-tuned} & RoBERTa   & \textbf{.80} \\
        \textit{fine-tuned} & SciBERT & \textbf{.76} \\
        \bottomrule
        \end{tabular}
        \end{threeparttable}
    \end{minipage}%
    \hspace{0.05\linewidth}
    \begin{minipage}[t]{.47\linewidth}
      \centering
        \footnotesize
        \caption{F1-Macro scores of BART and RoBERTa on the COVID-CQ dataset using different \textbf{P}re-\textbf{T}raining and \textbf{F}ine-\textbf{T}uning Tokenizers. All models were pre-trained on the CORD-19 dataset.}\label{tab:tokenizer_inv}
        \vspace{3.9mm}
        \begin{threeparttable}
        \begin{tabular}{lccr}\toprule
        \textbf{Models} & \textbf{PT Tok.} & \textbf{FT Tok.}  & \textbf{F1-Macro}     \\ \midrule
        RoBERTa	& Standard & Standard	& \textbf{.78}\\
        RoBERTa	& COVID & Standard	& .73\\
        RoBERTa	& COVID & COVID	& .72\\
        BART  & Standard & Standard	& \textbf{.77}\\
        BART  & COVID & Standard	& .73\\
        BART  & COVID & COVID	& .70\\
        \bottomrule
        \end{tabular}
        \end{threeparttable}
    \end{minipage} 
\end{table*}

\begin{table*}[!tb]
\centering
\caption{Average F1-Macro scores and standard deviation over three randomly sampled runs of neural language models for COVID datasets. The table is divided into three parts: general-purpose baselines, intermediate pre-trained, and COVID-19 intermediate pre-trained models. \textbf{C}OVID \textbf{A}ware means the model was pre-trained on CORD-19 (\ding{51}), pre-trained on a different dataset (\ding{55}), or the dataset was not reported (\textbf{?}). \textbf{I}ntermediate \textbf{T}raining means the model was pre-trained in a specific domain.
\textbf{Boldface} indicates the highest value for each dataset. $^\dagger$Models trained on CORD-19. $^*$Large version of the model.} \label{tab:fine_tuned}
\begin{threeparttable}
\resizebox{\linewidth}{!}{
\begin{tabular}{cclccccc}\toprule
\textbf{IT} & \textbf{CA} & \textbf{Model} & \textbf{CMU-MisCov19} & \textbf{CoAID} & \textbf{ReCOVery} & \textbf{COVID19FN} & \textbf{COVID-CQ}     \\
\midrule

- & - & BERT	&  .54$\pm$.03 &  .93$\pm$.01 &  .78$\pm$.02 &  .65$\pm$.01 &  .76$\pm$.01	  \\
- & - & RoBERTa &  .53$\pm$.03 &   .95$\pm$.01 &    .81$\pm$.01 &   .73$\pm$.02 &  .64$\pm$.01  \\
- & - & BART	&  .49$\pm$.03 &   \textbf{.96}$\pm$.01 &  .90$\pm$.01 &  \textbf{.83}$\pm$.01 &  .75$\pm$.01	 \\  
- & - & DeBERTa &  .52$\pm$.04 &  .95$\pm$.01 &   .75$\pm$.01 &  .67$\pm$.03 &  .63$\pm$.02	\\ \hdashline\noalign{\vskip .5ex}

\ding{51} & - & SciBERT &  .46$\pm$.02 &  .95$\pm$.01 &   .77$\pm$.01 &  .76$\pm$.01 &  .61$\pm$.02	\\
\ding{51} & - & BioClinicalBERT	&  .48$\pm$.02 &  .89$\pm$.01 &  .82$\pm$.01 &  .81$\pm$.02 &  .63$\pm$.02 \\
\ding{51} & - & BERTweet &  .51$\pm$.03 &  .88$\pm$.02 &  .84$\pm$.03 &   .65$\pm$.01 &  .75$\pm$.01 \\
\hdashline\noalign{\vskip .5ex}

\ding{51} & \ding{55} & CT-BERT $^*$ &  \textbf{.58}$\pm$.04 &  .94$\pm$.01 &  .80$\pm$.02 &      .40$\pm$.01 &   .63$\pm$.02 \\
\ding{51} & \textbf{?} & ClinicalCOVID-BERT	&  .50$\pm$.03 &  .93$\pm$.01 &  .82$\pm$.01 &  .78$\pm$.02 &  .74$\pm$.02\\ 
\ding{51} & \textbf{?} & BioCOVID-BERT $^*$	&    .45$\pm$.01 &  .91$\pm$.02 &    .81$\pm$.01 &  .68$\pm$.03 &  .63$\pm$.02 \\

\ding{51} & \ding{51} & COVID-BERT $^\dagger$	&  .46$\pm$.02 &  .94$\pm$.01 &  .85$\pm$.03 &  .73$\pm$.02 &  .72$\pm$.01 \\
\ding{51} & \ding{51} & COVID-SciBERT $^\dagger$ & .34$\pm$.02	& .92$\pm$.03	& .78$\pm$.03	& .67$\pm$.02	& .76$\pm$.03 \\
\ding{51} & \ding{51} & COVID-RoBERTa	& .40$\pm$.04	& .92$\pm$.04	& \textbf{.91}$\pm$.03	& .67$\pm$.03	& \textbf{.78}$\pm$.03 \\
\ding{51} & \ding{51} & COVID-BART	& .33$\pm$.01	& .91$\pm$.06	& .89$\pm$.03	& .66$\pm$.05	& .77$\pm$.07 \\
\ding{51} & \ding{51} & COVID-DeBERTa	& .30$\pm$.01	& .92$\pm$.05	& .89$\pm$.02	& .81$\pm$.03	& .73$\pm$.01 \\

\bottomrule
\end{tabular}
}
\end{threeparttable}
\end{table*} 

All models achieved low scores for CMU-MisCov19, making it the most challenging dataset in our evaluation. The best results were obtained for CoAID. Overall, general-purpose baselines achieved comparable results to COVID-19 intermediate pre-trained models for all datasets. For example, the best mean result for the dataset ReCOVery was achieved by COVID-RoBERTa (F1=.91, std=.03) while the general-purpose model BART (F1=.90, std=.01) was only .01 score points worse. We observe similar results for COVID-CQ, where the best model COVID-RoBERTa (F1=.78, std=.03) has a score difference of .02 to the second-best model BERT (F1=.76, std=.01).
We conclude that pre-training language models on COVID data before fine-tuning on a misinformation task did not generally provide an advantage for the tested datasets in this paper. 

\section{Conclusion \& Future Work}
This study empirically evaluated 15 Transformer models for five COVID-19 misinformation tasks. Our analysis shows domain-specific models and tokenizers do not generally perform better in the classification of misinformation. We conclude that the vocabulary related to COVID-19 and possibly text-patterns about COVID-19 do not have a significant effect on the models' ability to classify misinformation. 

The main limitation of our study is the non-standardized pre-training of models due to the models' diversity.
To reliably detect misinformation across content types and platforms, researchers need access to diverse data. We see this study as an initial step to compile a benchmark for COVID-19 data similar to widely adopted natural language understanding benchmarks (e.g., GLUE, SuperGLUE) which enable an evaluation across diverse sets of misinformation domains, sources, and tasks.

Controlling the current and future pandemics requires reliable detection of false information propagated through many streams and having different unique features.
This study is a first step for researchers and policymakers to devise and deploy systems that reliably flag misinformation related to COVID-19 from a broad spectrum of sources.

The usefulness of NLP models increases significantly if they are applicable to multiple tasks \cite{WangPNS19}. We anticipate future NLP technologies for detecting misinformation need to adopt the trend of evaluating on several benchmark datasets. This work provides a first milestone in evaluating general model capabilities and questioning the advantage of domain-specific model pre-training.

Although COVID-19 accelerated the propagation of misinformation and disinformation, these problems are not unique to the current pandemic. The effects of COVID-19 misinformation and disinformation on elections, ethical biases, and the portrayal of ethical groups \cite{BenklerFR18} can have similar or even more severe consequences on society than misinformation related to COVID-19. Therefore, identifying false information streams across domains will remain a challenging problem, and identifying which models can generalize for many sources is crucial.

\bibliographystyle{splncs04}
\bibliography{mybib}

\clearpage

\appendix

\section{Appendix}
\label{sec:appendix}

\subsection{Dataset Details}
\label{sec:appendix_dataset_details}

\textbf{COVID-19 Open Research Dataset} (CORD-19)~\cite{WangLCR20} is the largest open source dataset about COVID-19 and coronavirus-related research (e.g. SARS, MERS). CORD-19 is composed of more than 280K scholarly articles from PubMed,\footnote{\url{https://pubmed.ncbi.nlm.nih.gov/}} bioRxiv,\footnote{\url{https://www.biorxiv.org/}} medRxiv,\footnote{\url{https://www.medrxiv.org/}} and other resources maintained by the WHO.\footnote{\url{https://www.who.int/}} We use this dataset to extend the general pre-training from selected neural language models (cf. \Cref{sec:methodology}) into the COVID-specific vocabulary and features.

\textbf{Covid-19 heAlthcare mIsinformation Dataset} (CoAID)~\cite{CuiL20} focuses on healthcare misinformation, including fake news on websites, user engagement, and social media. CoAID is composed of 5\,216 news articles, 296\,752 related user engagements, and 958 posts about COVID-19, which are broadly categorized under the labels \textit{true} and \textit{false}.

\textbf{Twitter Stance Dataset} (COVID-CQ)~\cite{MutluOJT20} is a dataset of user-generated Twitter content in the context of COVID-19. More than 14K tweets were processed and annotated regarding the use of \textit{Chloroquine} and \textit{Hydroxychloroquine} as a valid treatment or prevention against the coronavirus. COVID-CQ is composed of 14\,374 tweets from 11\,552 unique users labeled as \textit{neutral}, \textit{against}, or \textit{favor}.

\textbf{ReCOVery}~\cite{ZhouMFZ20} explores the low credibility of information on COVID-19 (e.g., bleach can prevent COVID-19) by allowing a multimodal investigation of news and their spread on social media. 
The dataset is composed of 2\,029 news articles on the coronavirus and 140\,820 related tweets labeled as \textit{reliable} or \textit{unreliable}.  

\textbf{CMU-MisCov19}~\cite{MemonC20} is a Twitter dataset created by collecting posts from unknowingly misinformed users, users who actively spread misinformation, and users who disseminate facts or call out misinformation. CMU-MisCov19 is composed of 4\,573 annotated tweets divided into 17 classes (e.g., \textit{conspiracy}, \textit{fake cure}, \textit{news}, \textit{sarcasm}). The high number of classes and their imbalanced distribution make CMU-MisCov19 a challenging dataset.

\textbf{COVID19FN}\footnote{\url{https://tinyurl.com/4mryzj5k}} is composed of approximately 2\,800 news articles extracted mainly from Poynter\footnote{\url{https://www.poynter.org/ifcn/}} categorized as either \textit{real} or \textit{fake}.

\subsection{Model Details}
\label{sec:appendix_model_details}

\textbf{General-Purpose Baselines.} BERT~\cite{DevlinCLT19} mainly captures general language characteristics using a bidirectional \textit{Masked Language Model} (MLM) and \textit{Next Sentence Prediction} (NSP) tasks.
RoBERTa~\cite{LiuOGD19} improves BERT with additional data, compute budgets, and hyperparameter optimizations. RoBERTa also drops the NSP as it contributes little to the model representation.  
BART~\cite{LewisLGG19} optimizes an auto-regressive forward-product and auto-encoding MLM objective simultaneously.
DeBERTa~\cite{HeLGC21} improves the attention mechanism using a disentanglement of content and position. 

\textbf{Intermediate Pre-Trained.} SciBERT~\cite{BeltagyLC19a} optimizes the MLM for 1.14M randomly selected papers from Semantic Scholar\footnote{\url{https://www.semanticscholar.org/}}. BioClinicalBERT~\cite{AlsentzerMBW19} specializes on 2M notes in the MIMIC-III database~\cite{mimiciii}, a collection of disidentified clinical data. BERTweet \cite{NguyenVT20} optimizes BERT on 850M tweets each containing between 10 and 64 tokens.

\textbf{COVID-19 Intermediate Pre-Trained}
COVID-Twitter-BERT~\cite{MullerSK20} (CT-BERT) uses a corpus of 160M tweets for domain-specific pre-training and evaluates the resulting model's capabilities in sentiment analysis, such as for tweets about vaccines. BioClinicalBERT~\cite{AlsentzerMBW19} fine-tunes BioBERT~\cite{LeeYKK19} into clinical narratives in the hope to incorporate linguistic characteristics from both the clinical and biomedical domains.

Cui et al. \cite{CuiL20} propose CoAID and investigate the misinformation detection task by comparing traditional machine learning (e.g., logistic regression, random forest) and deep learning techniques (e.g., GRU). In a similar layout, Zhou et al. \cite{ZhouMFZ20} compare traditional statistical learners, such as SVM and neural networks (e.g., CNN), to classify news as credible or not. In both studies, the results show deep learning architectures as the most prominent options.


\subsection{Evaluation Details}

\textbf{Pre-Training.} We use the data from the abstracts of the CORD-19 dataset for pre-training. For pre-processing the CORD-19 abstract data, we consider only alphanumerical characters. We use a sequence length of 128 tokens, which reduces training time while being competitive to longer sequence lengths when fine-tuning \cite{Press20}. We mask words randomly with a probability of .15, a common configuration for Transformers \cite{DevlinCLT19,HeLGC21}, and perform the MLM with the following remaining parameters: a batch size of 16 for all the base models, and eight for the large models, the Adam Optimizer ($\alpha = 2e-5$, $\beta_1 = .9$, $\beta_2 = .999$, $\epsilon = 1e-8$), and a maximum of five epochs. All experiments were performed on a single NVIDIA GeForce GTX 1080 Ti GPU with 11 GB of memory.

\textbf{Fine-Tuning.} The classification model applies a randomly initialized fully-connected layer to the aggregate representation of the underlying Transformer (e.g., \texttt{[CLS]} for BERT) with dropout ($p=.1$) to learn the annotated target classes with cross-entropy loss for five epochs and with a sequence length of 200 tokens. We use the same configuration of the optimizer as in pre-training.

\end{document}